\documentclass[final]{cvpr}

\usepackage{times}
\usepackage{epsfig}
\usepackage{graphicx}
\usepackage{amsmath}
\usepackage{amssymb}

\usepackage[pagebackref,breaklinks,colorlinks]{hyperref}

\def\cvprPaperID{6361} 
\def\confName{ICCV}
\def\confYear{2021}

\usepackage{booktabs}
\usepackage{multirow} 
\usepackage{color}

\usepackage[accsupp]{axessibility}

\begin{document}

\title{
UltraPose: Synthesizing Dense Pose with 1 Billion Points by Human-body Decoupling 3D Model
}


\author{Haonan Yan$^{1\ast}$, ~~Jiaqi Chen$^{2}$\thanks{~Equal contribution. ~$\dag$ Corresponding authors.}~, ~~Xujie Zhang$^{2}$, ~~Shengkai Zhang$^{1}$, \\
~~Nianhong Jiao$^{1}$, ~~Xiaodan Liang$^{2\dag}$,  ~~Tianxiang Zheng$^{1\dag}$ \\
{$^1$Beijing Momo Technology Co., Ltd. ~~~$^2$Sun Yat-sen University} \\
{\small\texttt{yan.haonan@immomo.com, \{jadgechen,xdliang328,zhengtianxiang1128\}@gmail.com, }}\\
{\small\texttt{zhangxj59@mail2.sysu.edu.cn, songkey@pku.edu.cn, jnhrhythm@tju.edu.cn}}
}

\maketitle
\pagestyle{empty}  
\thispagestyle{empty} 

\begin{abstract}
Recovering dense human poses from images plays a critical role in establishing an image-to-surface correspondence between RGB images and the 3D surface of the human body, serving the foundation of rich real-world applications, such as virtual humans, monocular-to-3d reconstruction. However, the popular DensePose-COCO dataset relies on a sophisticated manual annotation system, leading to severe limitations in acquiring the denser and more accurate annotated pose resources. In this work, we introduce a new 3D human-body model with a series of decoupled parameters that could freely control the generation of the body. Furthermore, we build a data generation system based on this decoupling 3D model, and construct an ultra dense synthetic benchmark \textbf{UltraPose}, containing around 1.3 billion corresponding points. Compared to the existing manually annotated DensePose-COCO dataset, the synthetic UltraPose has ultra dense image-to-surface correspondences without annotation cost and error. Our proposed UltraPose provides the largest benchmark and data resources for lifting the model capability in predicting more accurate dense poses. To promote future researches in this field, we also propose a transformer-based method to model the dense correspondence between 2D and 3D worlds. The proposed model trained on synthetic UltraPose can be applied to real-world scenarios, indicating the effectiveness of our benchmark and model.\footnote{~Dataset and code: \href{https://github.com/MomoAILab/ultrapose}{https://github.com/MomoAILab/ultrapose}}

\end{abstract}


\begin{figure*}[t]
\begin{center}
 \includegraphics[width=0.96\textwidth]{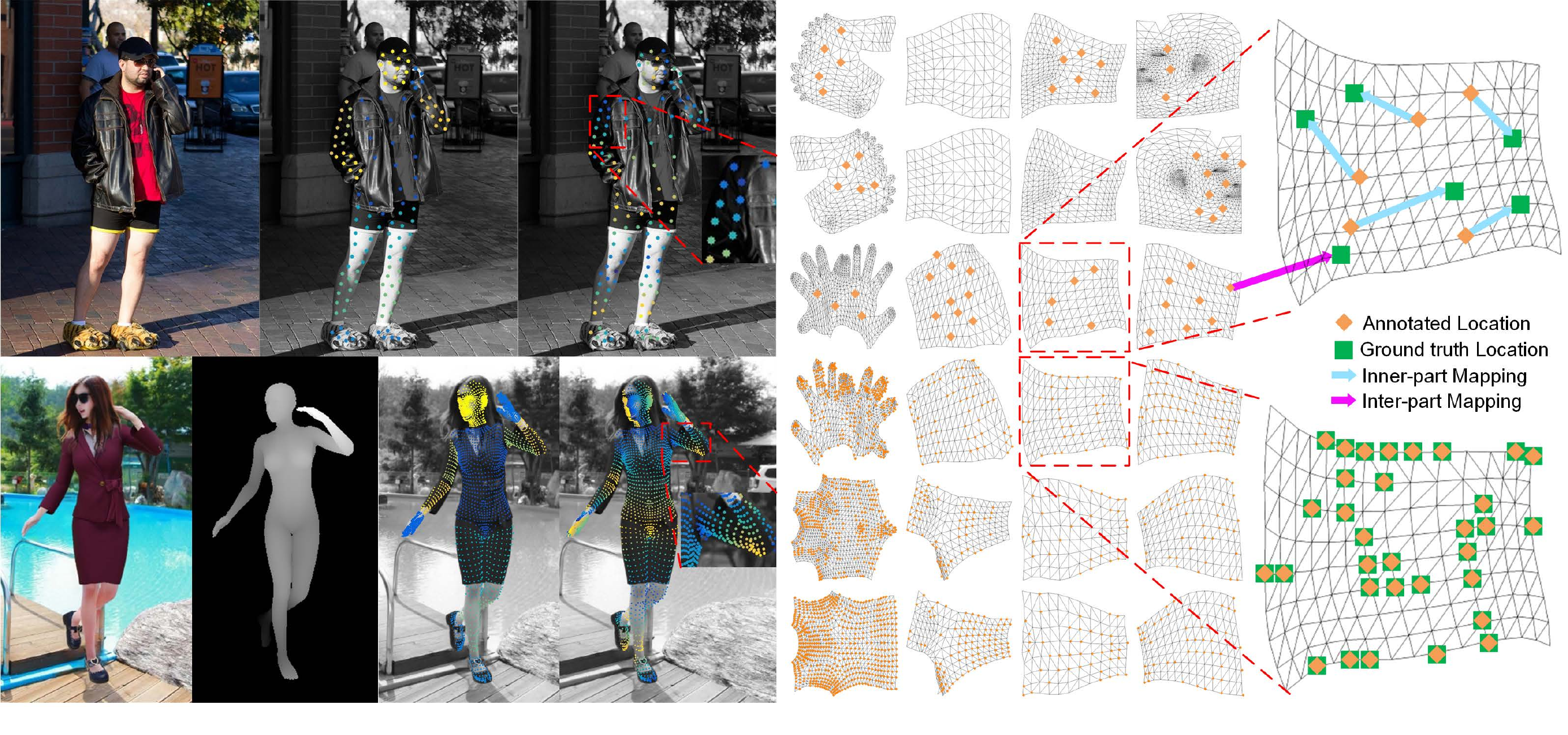}
 \vspace{-5mm}
\end{center}
  \caption{Visualization comparison of annotations on DensePose-COCO and our proposed UltraPose benchmark. The DensePose-COCO data (above) has relatively sparse point annotations, while our UltraPose (below) has ultra dense annotations, providing a depth map of the human body simultaneously. 
  Besides, we also show their UV space comparison of annotation on the right, in which DensePose-COCO data has obvious annotation errors. For instance, there are many inner-part (blue lines) errors and inter-part (purple lines) errors. 
  However, our UltraPose is generated with an error-free label that all the annotated and ground-truth points coincide perfectly.
  }
\label{fig:dataset}
\end{figure*}

\section{Introduction}
In computer vision, establishing correspondences from 2D images to 3D human body models is a fundamental task for analyzing human action, which facilitates several application scenarios, such as texture transfer~\cite{neverova2018dense,pavlakos2019texturepose,grigorev2019coordinate}, virtual try-on~\cite{neuberger2020image,kubo2019uvton,wu2019m2e,xie2021vton}, and 3D human reconstruction~\cite{guler2019holopose,weng2019photo}.
Based on the Skinned Multi-Person Linear (SMPL) model~\cite{loper2015smpl}, DensePose~\cite{guler2018densepose} takes a single RGB image as input and maps pixels to corresponding surface points on the SMPL model, obtaining more accurate instance-level human analysis with these predicted denser relationships.
Recently, several methods~\cite{yang2019parsing,neverova2019correlated,neverova2019slim} have been proposed for dense pose estimation, while the results show that it is still a very challenging problem.

One of the limitations of the current DensePose system is that the SMPL is a model for academic research and is not widely recognized in the industry, which limits its  application value in real-world scenarios. As a skinned model that can represent a wide variety of body shapes, SMPL has only 10 blend shapes that do not have a clear physical meaning and may affect each other. Therefore, some problems arise when applying SMPL model, such as poor controllability which makes it difficult to adjust the human body into the targeted shape.

There are also some limitations in the popular DensePose-COCO dataset~\cite{guler2018densepose}. On the one hand, it only collects 50K persons, not covering various poses and shapes, which makes the model perform poorly in situations such as sideways and occlusion. On the other hand, the quality of manual annotations is intrinsically limited as described in ~\cite{neverova2019correlated}. Under some ambiguity and self-occlusion conditions, the annotators need to mark the exact point correspondences, resulting in huge annotation errors. 

In this work, to address the limitations of SMPL, we adopt an industry-recognized Daz\footnote{Daz: \href{https://www.daz3d.com/}{https://www.daz3d.com/}} model as the base human body model. Then, we propose a new 3D model, DeepDaz, containing a group of well-designed decoupling parameters that can control the generation of a variety of human bodies. These parameters have a specific physical meaning and are decoupled with each other, enabling humans to adjust the human body freely rather than relying on professional CG design. 
Our DeepDaz model is also compatible with the CG industry that can be freely edited in mainstream design software, thus leading to an excellent performance and application value.

Based on the DeepDaz model, we build a data generation system and further propose an ultra dense synthetic dataset, UltraPose, which contains 500K persons and 1.3B corresponding point annotations on the surface of DeepDaz model. Figure~\ref{fig:dataset} show the comparison between DensePose-COCO dataset and our UltraPose, which has several appealing properties. 
\textbf{First}, UltraPose has an ultra dense annotation with around 2.6K points in one person (about 25 times of DensePose-COCO) for pose estimation, which can promote researches on instance-level human analysis.
\textbf{Second}, based on the established data generation system, we can acquire quantities of data with rich diversity and no manual annotation costs.   
\textbf{Third}, the generated data annotations are absolute truth values without any error, and perfectly represent the corresponding relationship between the 2D image and the surface of human body. 
\textbf{Fourth}, the UltraPose also provides the 3D parameters and depth information of the human body for further research.

Tackling such a large-scale and diverse benchmark is still challenging, requiring the model to have strong semantic feature representation capabilities.
Inspired by \cite{chen2021transunet}, we design a new transformer-based model for dense pose estimation task. 
It combines the merits of Transformers~\cite{vaswani2017attention} and U-Net~\cite{ronneberger2015u} simultaneously, and also uses the prior keypoints knowledge to assist in prediction. 
Our proposed model obtains state-of-the-art accuracy on the UltraPose benchmark, and more importantly, can be directly applied to real-world scenarios, achieving impressive performance. 

In summary, the major contributions of this work are three-fold:
\begin{itemize}
  \item We replace the SMPL model with DeepDaz, a new human body 3D decoupling model, which can be used to generate various poses easily and is compatible with CG industry design standards. 
  \item We propose a realistic human body generation system and a new large-scale synthesized benchmark, UltraPose, which contains 1.3 billion points annotation without any annotation cost or error.
  \item A transformer-based method that can extract informative visual representations for ultra dense pose estimation. After being trained on the UltraPose dataset, our proposed methods also can be applied to real-world dense pose estimation. 
\end{itemize}

\begin{figure*}[t]
\begin{center}
 \includegraphics[width=0.96\textwidth]{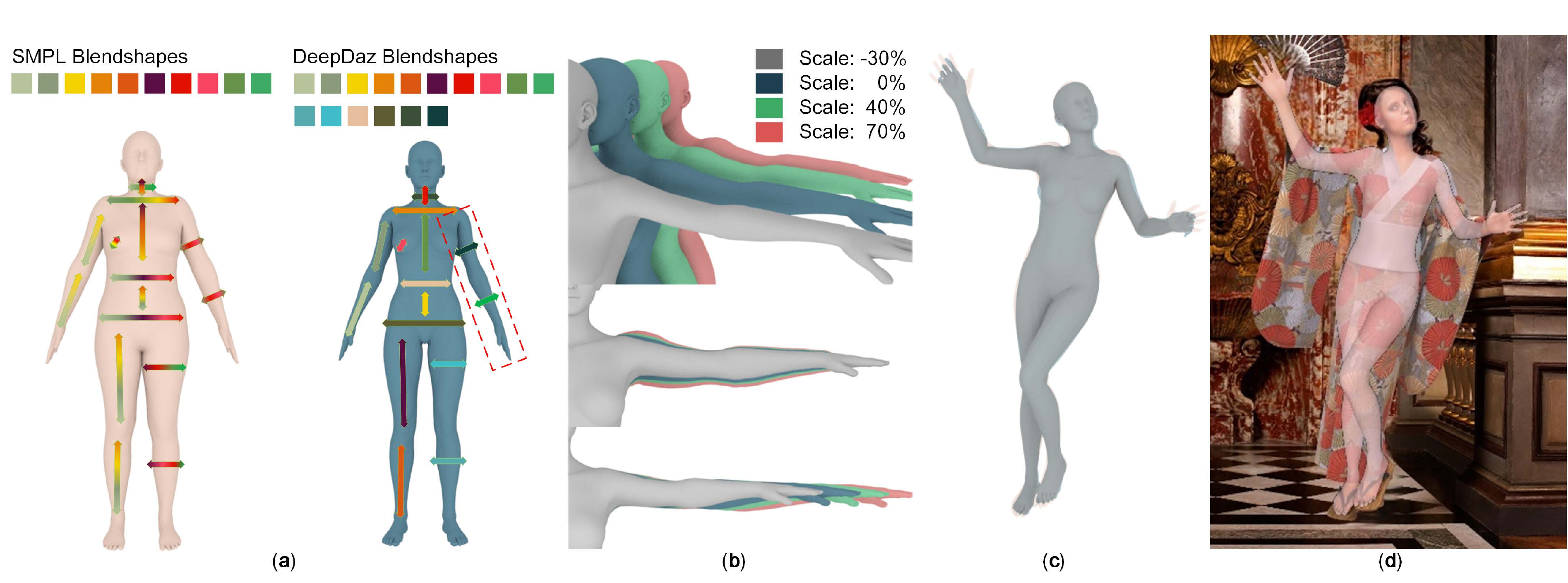}
 \vspace{-5mm}
\end{center}
  \caption{ (\textbf{a}) Comparison between SMPL shape components and DeepDaz shape components. (\textbf{b}) The influence of different parameter scales on arm shape. (\textbf{c}) Errors between DeepDaz model and SMPL model after using NICP point cloud registration. (\textbf{d}) Errors between SMPL model and synthetic image.
  }
\vspace{-0mm}
\label{fig:daz}
\end{figure*}

\section{Related Work}

\paragraph{Dense Pose Estimation}
Skinned Multi-Person Linear Model (SMPL)~\cite{loper2015smpl} is a learned model of human body, containing multiple controllable shape and pose parameters
Motivated by SMPL and DenseReg~\cite{alp2017densereg}, Guler \textit{et al.}~\cite{guler2018densepose} model the 3D body structure in a 2D image and introduce a new benchmark DensePose-COCO that presents the dense correspondences between images and the SMPL model. Based on Mask-RCNN~\cite{he2017mask}, Guler \textit{et al.}~\cite{guler2018densepose} also propose a dense regression framework, DensePose-RCNN, to predict the body part segmentation and UV coordinates.
Several other methods~\cite{yang2019parsing,neverova2019correlated,neverova2019slim,neverova2020continuous,sanakoyeu2020transferring} are proposed to address the challenging dense pose estimation. 
Yang \textit{et al.} present Parsing R-CNN~\cite{yang2019parsing}, a flexible and efficient pipeline that adopts FPN~\cite{lin2017feature} backbone and RoIAlign~\cite{he2017mask} operation. Parsing R-CNN is the champion in the
COCO 2018 Challenge DensePose Estimation task and can be applied to multiple instance-level human analysis tasks, such as human part segmentation. 
Neverova \textit{et al.}~\cite{neverova2019correlated} notice the quality limitation of manual annotations. Therefore, they propose an augmenting neural network that could predict the distribution over labeling data, thus understanding the annotation uncertainty better and archiving state-of-the-art accuracy. 
However, limited by the current dataset annotation quality and model design defects, the performance of these models is not satisfactory. Dense pose estimation is yet a challenging task.

\paragraph{Synthetic Datasets}
Recently, large-scale datasets have promoted the development of computer vision, such as ImageNet~\cite{deng2009imagenet}, MSCOCO~\cite{lin2014microsoft}, etc. However, labeling these datasets manually requires a huge annotation cost with inevitable errors. 
The collection of real human datasets is even harder, considering the privacy issues and the more complex labeling process~\cite{nikolenko2019synthetic}.
Fortunately, we have an alternative: synthetic dataset, which refers to building some specialized data generation systems to synthesize data that is as realistic as possible. 
Numerous works~\cite{rematas2018soccer,fabbri2018learning,varol17_surreal,sun2019dissecting,bak2018domain} have been proposed for synthesizing human body.
Varol \textit{et al.} present SURREAL~\cite{varol17_surreal}, a synthetic large-scale benchmark for 3D pose keypoints, depth map, and segmentation. SURREAL is rendered from 3D sequences of human motion that contains more than 6 million frames. 
To address the scarcity of human tracking, body part, and occlusion annotations, Fabbri \textit{et al.} propose JTA dataset~\cite{fabbri2018learning}, which is created by exploiting the highly photorealistic video game Grand Theft Auto V.
Dense pose estimation suffers a similar problem. Therefore, motivated by these works, we propose a new human body model DeepDaz, further construct a large-scale synthetic benchmark UltraPose.


\begin{figure*}[t]
\begin{center}
 \includegraphics[width=0.96\textwidth]{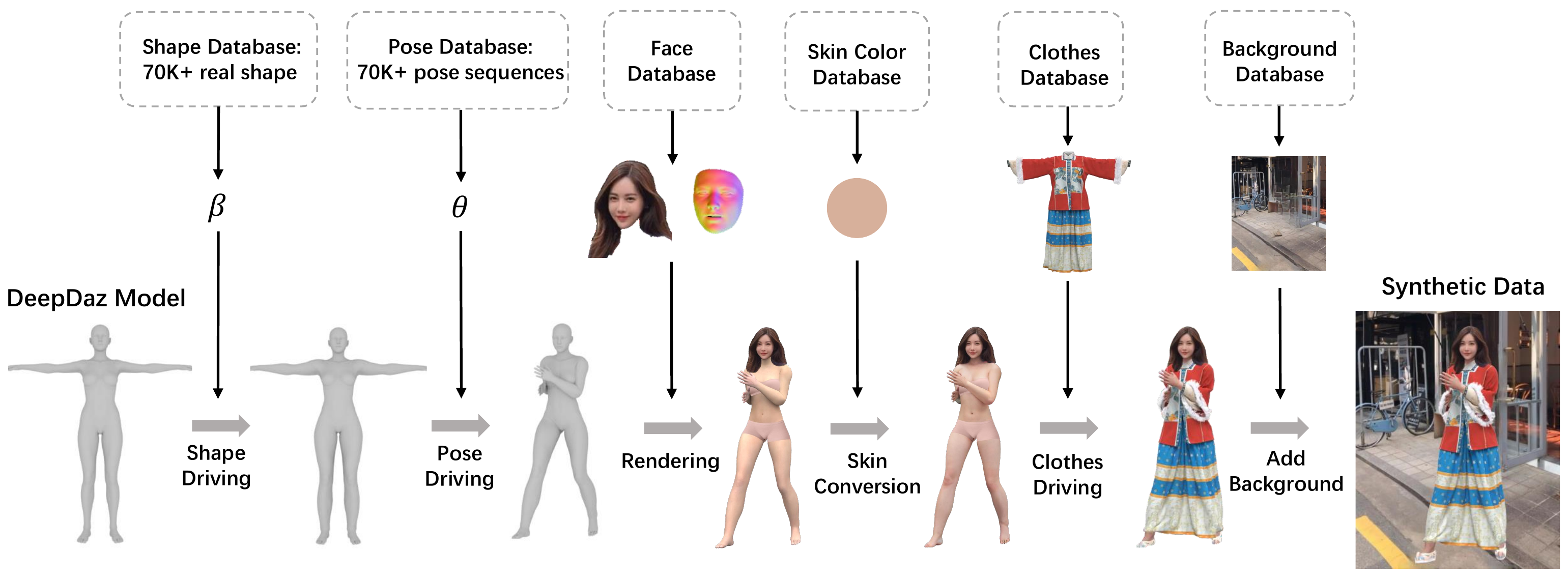}
 \vspace{-5mm}
\end{center}
  \caption{Data generation system. We randomly select various fundamental materials from the databases and get the final synthetic image data by the generation pipeline. Our generated data is sufficient fidelity for dense pose estimation.
  }
\vspace{-0mm}
\label{fig:generation}
\end{figure*}

\begin{table*}[t]
\centering
\renewcommand\tabcolsep{5.0pt}
\vspace{2mm}
\caption{ Comparison between our UltraPose and existing DensePose-COCO.}
\vspace{2mm}
\begin{tabular}{{l}*{7}{c}}
    \toprule	
    Dataset  & Persons & Points & \#Avg Density & Mask Resolution & No Error & Depth & 3D Parameters  \\ 
    \midrule
    \text{DensePose-COCO} \cite{guler2018densepose}  & 49K & 5.2M & 106 & 256×256 & & \\
        \textbf{UltraPose-5K (Ours)}  & 5k & 13M & 2.6K & 512×512 & \checkmark & \checkmark & \checkmark  \\
        \textbf{UltraPose (Ours)}  & 500K & 1.3B & 2.6K & 512×512 & \checkmark & \checkmark & \checkmark  \\
    \bottomrule
\end{tabular}

\label{table:dataset}
\end{table*}

\section{The UltraPose Dataset}
A large-scale high-quality dataset is crucial in modeling the correspondence between 2D images and 3D surface models. To address the limitations of the previous DensePose-COCO~\cite{guler2018densepose}, we introduce a new human body model (DeepDaz) and build UltraPose with 1.3 billion points, which is the first synthetic and the largest benchmark in dense pose estimation.
We explain the DeepDaz model, how to generate the dataset, and analyze the data in detail below.

\subsection{DeepDaz Model}
In this section, we explain the details of our proposed human body decoupling 3D model, DeepDaz.
The base model of DeepDaz is Daz, an art-designed human body that is widely recognized in the industry, composed of 16,556 vertexes, 32,736 surfaces, and 170 skeletons, including expressions, fingers, and toes joints.
For DeepDaz, we further design a series of freely adjustable decoupling parameters that can be controlled arbitrarily and produce a variety of human shapes.
The skeletons drive the model through the skinning algorithm to obtain various human bodies. Deepdaz is consistent with the human body driving standard of the CG industry and can be edited in design software, such as Maya, Blender, 3DMax, etc.

Although the driving algorithm is similar to the SMPL model, our DeepDaz has several notable advantages. \textbf{First}, DeepDaz provides decoupled shape components that can be controlled freely. As shown in Fig.~\ref{fig:daz} (a), the SMPL model contains 10 shape components that come from PCA of approximately 4000 scans, while DeepDaz currently contains 16 decoupled well-designed shape components. It's hard to generate a specific body shape using the statistical shape components in SMPL since people need to adjust several (usually 3-7) shape parameters simultaneously to get the expected body shape. In DeepDaz, people can simply adjust corresponding decoupled shape parameters to generate an expected body shape while keeping the other part shape unchanged (as shown in Fig.~\ref{fig:daz} (b)). \textbf{Second}, DeepDaz is extendible. People could design new shape components in CG software and integrate them into DeepDaz. \textbf{Third}, DeepDaz has better compatibility than the SMPL model. After generating the synthetic images, we can use NICP point cloud registration to translate DeepDaz parameters to the corresponding SMPL shape and pose parameters. The errors between SMPL, DeepDaz, and synthetic image are shown in Fig.~\ref{fig:daz} (c)(d).

\begin{figure*}[t]
\begin{center}
 \includegraphics[width=0.96\textwidth]{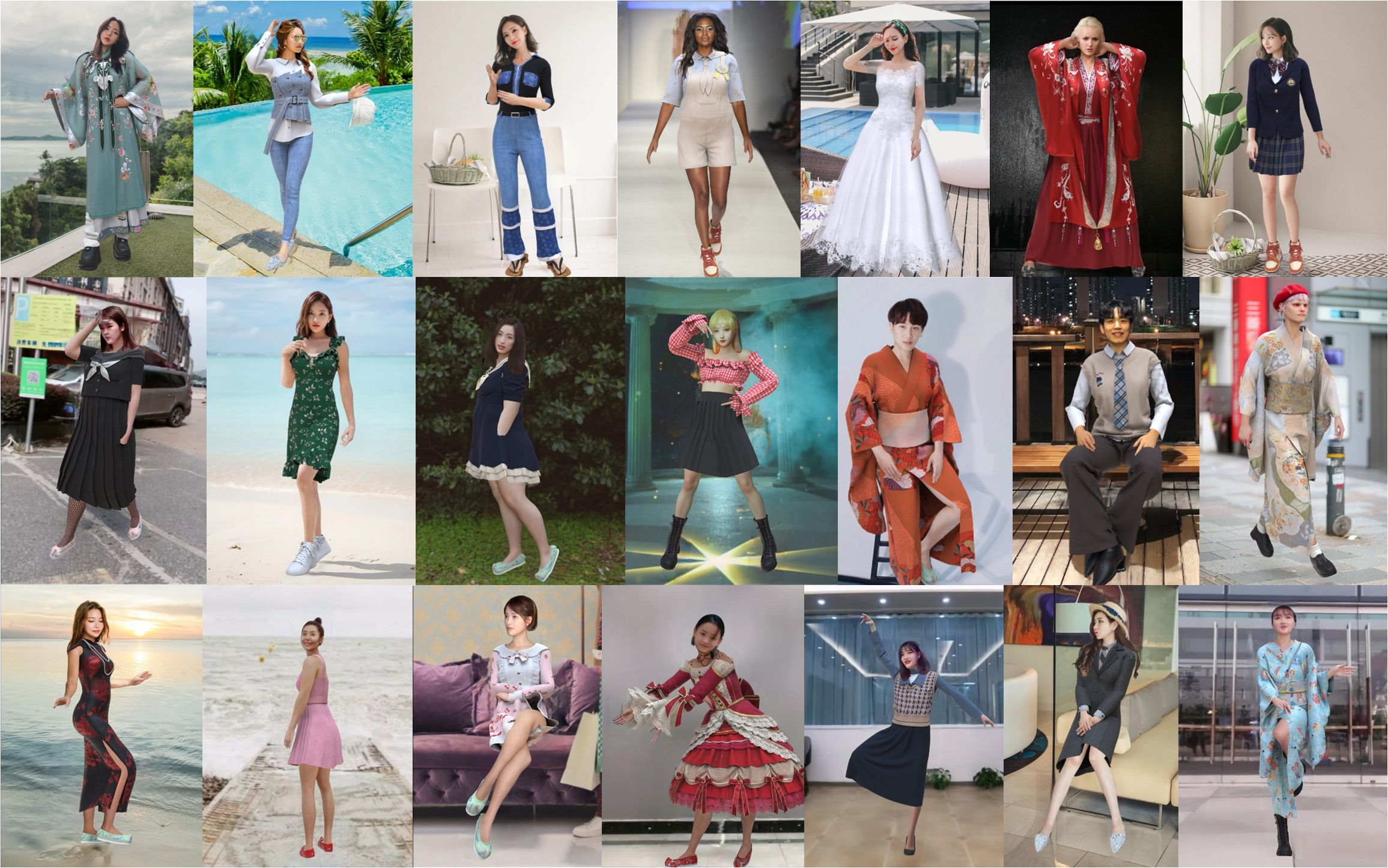}
 \vspace{-2mm}
\end{center}
  \caption{ Sample cases from our synthetic UltraPose benchmark with a large variety of poses, body shapes, clothings, heads, skin colors,  viewpoints and backgrounds.
  }
\label{fig:synthetic}
\end{figure*}

\subsection{Generation System}
We establish a data generation system that can quickly produce a large quantity of data. As shown in Fig.~\ref{fig:generation}, the system contains many fundamental material databases, such as shape database, pose database, clothes database, etc.
Furthermore, a data generation pipeline processes and merges the materials in the databases step by step and finally gets synthetic data.  
\vspace{-4mm}
\subsubsection{Databases Preparation} 
In detail, the fundamental material databases contain the shape database, pose database, face database, skin color database, clothes database, and background database. 
For the body shape, we sample 70K shapes from the real bodies to construct a shape database. 
To create a human pose database, we first collect 2,421 action sequences from open source databases such as Xbot, H36M~\cite{h36m_pami}, and SFU~\cite{sfu}, decomposing 70K poses from them. In addition, the human body will appear randomly in the 3D space within a certain range.
The face database contains 36K high-resolution front face images. 
The clothing database includes 374 pieces of clothes, shoes, and hats, which are covered on a skinned human model.
The background database has 65K background images that will be selected randomly.
These basic material databases ensure the diversity of the generated images.
\vspace{-4mm}
\subsubsection{Generation Pipeline}
\paragraph{Shape \& Pose Driving}
Similar to the SMPL model~\cite{loper2015smpl}, our DeepDaz also can generate target human body from parameters $\beta$ and $\theta$. These parameters are randomly selected from the shape and pose databases, controlling the generation of human body through the skinning algorithm.
\vspace{-4mm}
\paragraph{Rendering}
The rendering module performs data enhancement on the generated human body, including HDR, light angle, and light intensity. 
Motivated by 3DDFA parameters~\cite{3ddfa_cleardusk,guo2020towards} of the face and the GAN model, we also generate a realistic face that matches the head.
We finally obtain a rendered human with skin and face.
\vspace{-4mm}
\paragraph{Skin Conversion}
These rendered human data are still relatively close to the virtual characters in the game. Referring to pix2pix network~\cite{isola2017image}, we design the skin color conversion module to specifically process the human skin color, generating a more realistic human body.
\vspace{-4mm}
\paragraph{Clothes Driving}
When driving clothing, two methods are used: one is skin driving, and the other is cloth simulation driving that simulates physical collisions to get a more natural vertex position. We have optimized the effect of cloth simulation to make the clothing more realistic, and integrate these two parts into the clothes driving module.
\vspace{-4mm}
\paragraph{Add Background}
We have achieved complete synthetic human body data. In this step, we randomly select an image from the background database as the background for the final synthetic image. Our pipeline can generate massive amounts of data quickly.

\subsection{Data Analysis}
We build two versions of datasets: UltraPose-5K and UltraPose, including 5K and 500K images respectively. Both datasets are divided into the train, validation, and test splits with a ratio of 80\%:10\%:10\%.
As shown in Table~\ref{table:dataset}, compared with DensePose-COCO, the UltraPose has more annotated persons. Based on the DeepDaz model and data generation system, our UltraPose has ultra dense corresponding point annotations, containing a total of 1.3B points annotations and the average point annotation density in one person is 2.6K, which is 250 times and 25 times that of DensePose respectively. What's more, the segmentation mask resolution in UltraPose is 512×512 (256×256 in DensePose), which leads to a fine dense pose estimation result.
More importantly, unlike the human annotation with inevitable errors, the synthetic data has an error-free annotation. In addition, we also obtain the depth map and 3D parameters of the human body from the data generation system for future research.

Sufficient fidelity is also a fundamental requirement for synthetic datasets.
Figure \ref{fig:synthetic} show some cases from our UltraPose benchmark. 
Compared with a synthetic human dataset SURREAL~\cite{varol17_surreal}  for pose keypoints estimation, our UltraPose benefits from the well-designed data generation system, acquiring more realistic human body data.
All these advantages have a significant impact on improving the performance of dense pose estimation.

\begin{figure}[t]
\begin{center}
 \includegraphics[width=1.0\columnwidth]{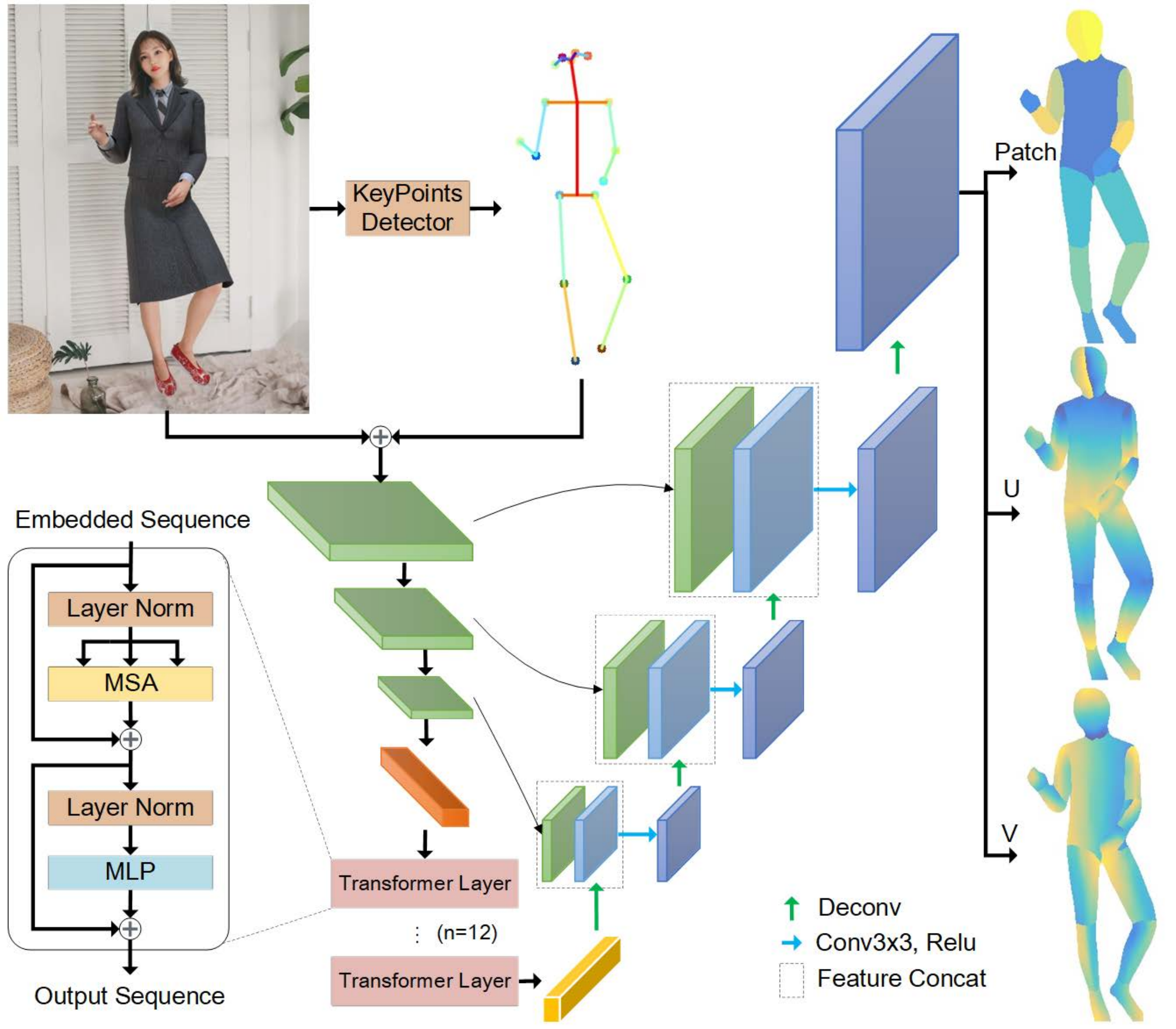}
 \vspace{-5mm}
\end{center}
  \caption{ Overview of the TransUltra architecture. TransUltra contains three main
components, including a keypoints detector, a U-Net framework, and multiple Transformer layers for representing image feature.
  }
\vspace{-0mm}
\label{fig:method}
\end{figure}

\section{Method}


In this work, we propose a transformer-based framework for dense pose estimation.
The overall architecture of the proposed TransUltra is shown in Figure~\ref{fig:method}.
We utilize the prior detected keypoints that are concatenated with the original image to guide the model to predict a more complete human body, which also helps solve the occlusion problem. 
Motivated by \cite{chen2021transunet}, we adopt some transformer~\cite{vaswani2017attention} layers to acquire informative feature for further representation. The U-Net architecture~\cite{ronneberger2015u} is applied in our framework since dense pose estimation is also a pixel-level prediction.

More specifically, we use OpenPose\footnote{\href{https://github.com/CMU-Perceptual-Computing-Lab/openpose}{https://github.com/CMU-Perceptual-Computing-Lab/openpose}} to obtain 17 key points which are formed into a keypoint image using visual post-processing. After concatenating with raw image, our network input is a 6-channel RGB-keypoints image of size
512×512. We use the first three stages of ResNet50 to extract a feature, which is then encoded into a feature sequence and fed to 12 transformer layers for further representation. Several deconvolution layers are also used to upsample the feature map back to the size of 512×512 and its channel size also decreases from 512 to 16 gradually. Finally, we apply a dense pose head to predict the body part segmentation and UV coordinates.

The model is trained within 80 epochs and optimized by Adam, during which the learning rate is set to 1e-4 and the batch size is 48.
Based on the synthetic UltraPose with perfect ground truth, we add some occlusions during training to simulate real-world scenes with strong occlusions. Benefiting from this, our model can address occluded human parts as shown in the first row of Figure~\ref{fig:vis}. However, occlusion prediction is not supported by the DensePose-COCO that doesn’t provide ground truth for occluded human parts.

\begin{table*}[t]
\centering
\caption{ Performance of different methods on the proposed UltraPose benchmark. Note that we use two versions (5K and all the data) of UltraPose to conduct experiments, which shows the effectiveness of a large-scale dataset.}
\vspace{3mm}
\begin{tabular}{{l}*{8}{c}}
    \toprule	
    Method & Data & ~~~~AP~~~~ & AP@GPS50 & AP@GPS75 & ~~AR~~ & AR@GPS50 & AR@GPS75\\ 
    \midrule
    Confidence Estimation \cite{neverova2019correlated} & 5K & 38.0 & 66.2 & 63.6 & 41.7 & 69.9 & 47.4 \\
    DensePose \cite{guler2018densepose} & 5K & 39.6 & 74.3 & 70.6 & 44.4 & 76.1 & 49.9 \\
    DeepLabV3 Head \cite{chen2017rethinking} & 5K & 43.1 & 75.9 & 74.9 & 47.5 & 78.8 & 55.3\\
    \textbf{TransUltra (Ours)} & 5K & 49.1 & 89.5 & 88.3 & 58.0 & 94.6 & 69.5\\
    \midrule
    Confidence Estimation \cite{neverova2019correlated} & 500K &  45.0 & 73.7  &  70.5 & 58.7 & 85.6  & 73.0  \\
    DensePose \cite{guler2018densepose} & 500K & 46.1 & 72.4  & 71.1 & 59.2  & 84.1  & 75.9  \\
    DeepLabV3 Head \cite{chen2017rethinking} & 500K & 52.3  & 81.9  & 80.9  &  63.3 & 90.1  & 80.8 \\
    \textbf{TransUltra (Ours)} & 500K & \textbf{56.2} & \textbf{91.8} & \textbf{91.8} & \textbf{63.6} & \textbf{95.9} & \textbf{84.7} \\
    \bottomrule
\end{tabular}

\label{table:UltraPose}
\end{table*}

\section{Experiments}
In this section, we first describe some related dense pose estimation baselines and the evaluation metric. We compare their performance with our proposed TransUltra model and discuss the gap between different methods. Finally, a qualitative result is shown to evaluate the generalization of TransUltra in real-world scenarios.

\subsection{Experimental Setup}
 
\paragraph{Baselines}

We implemented several existing methods as baselines. Their details are explained as follows.
DensePose~\cite{guler2018densepose}: the original DensePose-RCNN framework with fully convolutional head. 
DeepLabV3 Head~\cite{chen2017rethinking}: a powerful semantic image segmentation head with atrous convolution, which is applied to predict ultra dense correspondences between 2D image and the surface 3D human body model, since dense pose estimation is a pixel-level prediction task.
Confidence Estimation \cite{neverova2019correlated}: a network that models the correlated error fields for dense pose estimation. The quality limitation of manual annotations is discovered in this work, in which the model performs additional estimation of confidence in regressed UV coordinates, archiving state-of-the-art accuracy.

\paragraph{Evaluation Metric}
Following \cite{guler2018densepose}, we adopt geodesic point similarity (GPS) to measure the similarity of points prediction and ground truth annotation. 
The GPS metric is defined as below:
{
\setlength\abovedisplayskip{0.25mm}
\setlength\belowdisplayskip{0.25mm}
\begin{equation}
GPS_j = \frac{1}{|P_j|} \sum_{p \in P_j} exp \left( \frac{-g(i_p, \hat{i_p} ) ^2}{2k^2} \right)
\end{equation}
}%
where $P_j$ is the ground-truth annotations of j-th person, $i_p$ and $\hat{i_p}$ are the predicted  and ground-truth vertex respectively. $k$ is a normalizing parameter that is set  to 0.255.
After obtaining GPS similarity, we consider it as a threshold, and use the COCO challenge protocol to compute the Average Precision (AP) and Average Recall (AR). These metrics characterize the performance of a dense pose estimation algorithm.


\subsection{UltraPose Results}

We conduct extensive experiments on our UltraPose benchmark. It is worth noting that we adopt two versions of our dataset for experiments. One is the small version with 5K images, and the other contains all the data with 500K images and 1.3B annotated points.
We don’t provide quantitative experiments on the DensePose-COCO dataset, since our proposed UltraPose is a more advanced dataset with 1.3B error-free points annotation and realistic rendering. Besides, our model trained on the UltraPose will predict the occluded parts, which is not compatible with the DensePose-COCO that doesn’t provide occlusion annotation.
Experimental results in Table~\ref{table:UltraPose} show the comparison of methods and the impact of data quantity on the UltraPose dataset. 

\paragraph{Performance comparison}

On the complete UltraPose dataset, Confidence Estimation \cite{neverova2019correlated} archives an AP of 45.0 and the original DensePose-RCNN \cite{guler2018densepose} obtains 46.1 in the same metric. When applying the DeepLabV3 head \cite{chen2017rethinking}, we get a performance of 52.3 in AP metric. As a comparison, our proposed TransUltra obtains the best performance AP=56.2 by introducing the Transformer layer, U-Net framework, and pre-detected keypoints. The TransUltra also leads by a wide margin on other evaluation metrics, obtaining 91.8, 91.8, 63.6, 95.9, and 84.7 on the AP@GPS50, AP@GPS75, AR, AR@GPS50, and AP@GPS75 metrics, respectively.

\paragraph{The effectiveness of large-scale dataset}
We consider that the amount of data is crucial to the accuracy of the dense pose estimation. Experiments conducted on the UltraPose-5K dataset show that performance degrades significantly with only 5K data. For instance, on the AP metric, the performance of the TransUltra is 49.1, which is only 87.4\% of the performance obtained by training with 500K data.
This experiment also demonstrates the advantage of our synthetic data, since we can generate large amounts of error-free data without the need for manual annotation.
We find that 500K is a more appropriate amount of generated data and collect it as the UltraPose in this work.

\begin{figure*}[t]
\begin{center}
 \includegraphics[width=1.0\textwidth]{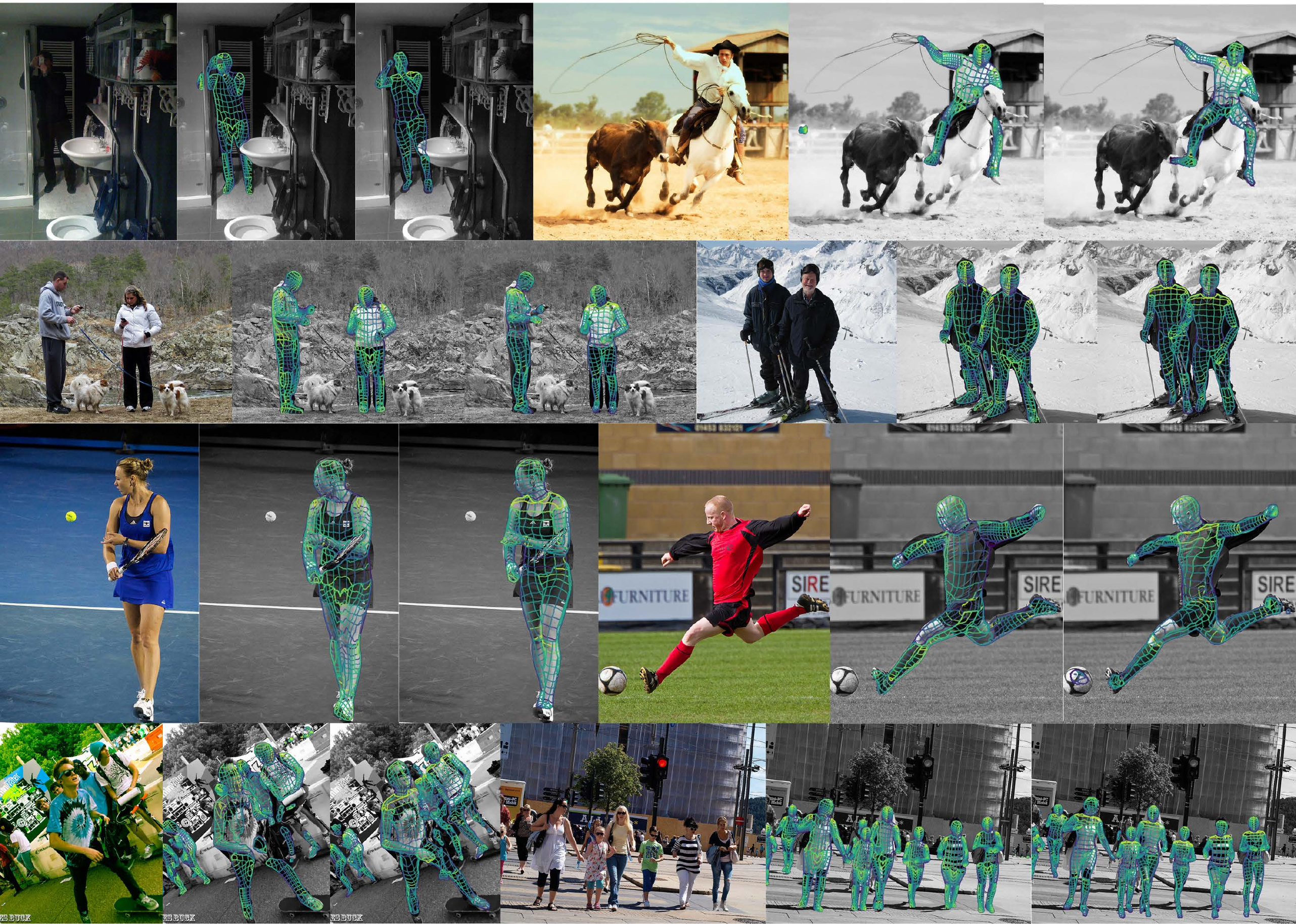}
\end{center}
  \caption{ Visualization comparison on DensePose-COCO dataset. Left: input image, Middle: the result of DensePose-RCNN, Right: the result of our proposed TransUltra model trained on synthetic UltraPose benchmark. These experiments have proved that our synthetic benchmark has sufficient fidelity, thus the model trained on UltraPose can be generalized to sophisticated real-world scenarios.
  }
\label{fig:vis}
\end{figure*}

\subsection{Qualitative Results}

In this section, we demonstrate the effectiveness of our proposed dataset and method.
We compare the results of two methods: (a) train and test the existing DensePose-RCNN on DensePose-COCO dataset, (b) train our proposed model on synthetic benchmark and test on real-world DensePose-COCO dataset.
In Figure \ref{fig:vis}, we show the visualization comparison results. The left is original input image, the middle is the DensPose-RCNN result, and the right is our model testing result on DensePose-COCO dataset.

Based on the synthetic UltraPose with perfect ground truth, we add some occlusions during training to simulate real-world scenes with strong occlusions. Benefiting from this, our model can address occluded human parts as shown in the first row of Figure \ref{fig:vis}. However, occlusion prediction is not supported by the DensePose-COCO that doesn’t provide ground truth for occluded human parts.

The second row shows the outstanding performance on handling clothes problem. Our data generation system first creates the human data with perfect ground truth, and then covers it with clothes. Therefore, using our synthetic data for training, the model can recover the body shapes under clothes more accurately.
However, the DensePose system manually annotates the body image covered with clothes, resulting in inevitable recovery errors in the model trained on DensePose-COCO.

The third row shows the results of sideways human body. For the DensePose-RCNN trained on DensePose-COCO, the result of sideways is relative worse that some incomplete predictions  occur. Benefits from utilizing the prior knowledge of human body keypoints, the proposed TransUltra handles the sideways situation well that the prediction result is complete and high-quality.

In the fourth row, we also demonstrate the model performance in some complex multi-person scenarios. These well-performed results prove that the proposed transformer-based model has a strong visual semantic representation ability and can be generalized to real-world scenarios in the DensePose-COCO dataset.

\section{Conclusion}

Modeling the correspondence between 2D image and 3D human body is a crucial task in computer vision, that can inspire large amounts of tasks, such as virtual try-on, 3D human reconstruction, etc.
In this work, we notice the limitations of the existing SMPL model and propose a new human-body decoupling model DeepDaz, which can be used to generate various human bodies. Then, we establish a data generation system, constructing the first large-scale ultra dense pose benchmark. Our UltraPose contains 1.3 billion point annotations without any error, and also provides the detailed parameters and depth map of the human body for further research. 
Finally, to facilitate future research in this field, we design a transformer-based model TransUltra, which is well trained on our UltraPose. The TransUltra not only achieves the best performance on our UltraPose benchmark, but also can be applied to real-world scenarios.

\paragraph{Acknowledgements}
This work was supported in part by National Key R\&D Program of China under Grant No.2020AAA0109700, National Natural Science Foundation of China (NSFC) under Grant No.U19A2073 and No.61976233, Guangdong Province Basic and Applied Basic Research (Regional Joint Fund-Key) Grant No.2019B1515120039, Guangdong Outstanding Youth Fund (Grant No.2021B1515020061), Shenzhen Fundamental Research Program (Project No.RCYX20200714114642083, No.JCYJ20190807154211365), Zhejiang Lab’s Open Fund (No.2020AA3AB14).

\clearpage

{\small
\bibliographystyle{ieee_fullname}
\bibliography{egbib}

\begin{thebibliography}{10}\itemsep=-1pt

\bibitem{alp2017densereg}
Riza Alp~Guler, George Trigeorgis, Epameinondas Antonakos, Patrick Snape,
  Stefanos Zafeiriou, and Iasonas Kokkinos.
\newblock Densereg: Fully convolutional dense shape regression in-the-wild.
\newblock In {\em Proceedings of the IEEE Conference on Computer Vision and
  Pattern Recognition}, pages 6799--6808, 2017.

\bibitem{bak2018domain}
Slawomir Bak, Peter Carr, and Jean-Francois Lalonde.
\newblock Domain adaptation through synthesis for unsupervised person
  re-identification.
\newblock In {\em Proceedings of the European Conference on Computer Vision
  (ECCV)}, pages 189--205, 2018.

\bibitem{chen2021transunet}
Jieneng Chen, Yongyi Lu, Qihang Yu, Xiangde Luo, Ehsan Adeli, Yan Wang, Le Lu,
  Alan~L Yuille, and Yuyin Zhou.
\newblock Transunet: Transformers make strong encoders for medical image
  segmentation.
\newblock {\em arXiv preprint arXiv:2102.04306}, 2021.

\bibitem{chen2017rethinking}
Liang-Chieh Chen, George Papandreou, Florian Schroff, and Hartwig Adam.
\newblock Rethinking atrous convolution for semantic image segmentation.
\newblock {\em arXiv preprint arXiv:1706.05587}, 2017.

\bibitem{deng2009imagenet}
Jia Deng, Wei Dong, Richard Socher, Li-Jia Li, Kai Li, and Li Fei-Fei.
\newblock Imagenet: A large-scale hierarchical image database.
\newblock In {\em 2009 IEEE conference on computer vision and pattern
  recognition}, pages 248--255. Ieee, 2009.

\bibitem{fabbri2018learning}
Matteo Fabbri, Fabio Lanzi, Simone Calderara, Andrea Palazzi, Roberto Vezzani,
  and Rita Cucchiara.
\newblock Learning to detect and track visible and occluded body joints in a
  virtual world.
\newblock In {\em European Conference on Computer Vision (ECCV)}, 2018.

\bibitem{grigorev2019coordinate}
Artur Grigorev, Artem Sevastopolsky, Alexander Vakhitov, and Victor Lempitsky.
\newblock Coordinate-based texture inpainting for pose-guided human image
  generation.
\newblock In {\em Proceedings of the IEEE/CVF Conference on Computer Vision and
  Pattern Recognition}, pages 12135--12144, 2019.

\bibitem{guler2019holopose}
Riza~Alp Guler and Iasonas Kokkinos.
\newblock Holopose: Holistic 3d human reconstruction in-the-wild.
\newblock In {\em Proceedings of the IEEE/CVF Conference on Computer Vision and
  Pattern Recognition}, pages 10884--10894, 2019.

\bibitem{guler2018densepose}
R{\i}za~Alp G{\"u}ler, Natalia Neverova, and Iasonas Kokkinos.
\newblock Densepose: Dense human pose estimation in the wild.
\newblock In {\em Proceedings of the IEEE conference on computer vision and
  pattern recognition}, pages 7297--7306, 2018.

\bibitem{3ddfa_cleardusk}
Jianzhu Guo, Xiangyu Zhu, and Zhen Lei.
\newblock 3ddfa.
\newblock \url{https://github.com/cleardusk/3DDFA}, 2018.

\bibitem{guo2020towards}
Jianzhu Guo, Xiangyu Zhu, Yang Yang, Fan Yang, Zhen Lei, and Stan~Z Li.
\newblock Towards fast, accurate and stable 3d dense face alignment.
\newblock In {\em Proceedings of the European Conference on Computer Vision
  (ECCV)}, 2020.

\bibitem{he2017mask}
Kaiming He, Georgia Gkioxari, Piotr Doll{\'a}r, and Ross Girshick.
\newblock Mask r-cnn.
\newblock In {\em Proceedings of the IEEE international conference on computer
  vision}, pages 2961--2969, 2017.

\bibitem{h36m_pami}
Catalin Ionescu, Dragos Papava, Vlad Olaru, and Cristian Sminchisescu.
\newblock Human3.6m: Large scale datasets and predictive methods for 3d human
  sensing in natural environments.
\newblock {\em IEEE Transactions on Pattern Analysis and Machine Intelligence},
  36(7):1325--1339, jul 2014.

\bibitem{isola2017image}
Phillip Isola, Jun-Yan Zhu, Tinghui Zhou, and Alexei~A Efros.
\newblock Image-to-image translation with conditional adversarial networks.
\newblock In {\em Proceedings of the IEEE conference on computer vision and
  pattern recognition}, pages 1125--1134, 2017.

\bibitem{kubo2019uvton}
Shizuma Kubo, Yusuke Iwasawa, Masahiro Suzuki, and Yutaka Matsuo.
\newblock Uvton: Uv mapping to consider the 3d structure of a human in
  image-based virtual try-on network.
\newblock In {\em Proceedings of the IEEE/CVF International Conference on
  Computer Vision Workshops}, pages 0--0, 2019.

\bibitem{lin2017feature}
Tsung-Yi Lin, Piotr Doll{\'a}r, Ross Girshick, Kaiming He, Bharath Hariharan,
  and Serge Belongie.
\newblock Feature pyramid networks for object detection.
\newblock In {\em Proceedings of the IEEE conference on computer vision and
  pattern recognition}, pages 2117--2125, 2017.

\bibitem{lin2014microsoft}
Tsung-Yi Lin, Michael Maire, Serge Belongie, James Hays, Pietro Perona, Deva
  Ramanan, Piotr Doll{\'a}r, and C~Lawrence Zitnick.
\newblock Microsoft coco: Common objects in context.
\newblock In {\em European conference on computer vision}, pages 740--755.
  Springer, 2014.

\bibitem{loper2015smpl}
Matthew Loper, Naureen Mahmood, Javier Romero, Gerard Pons-Moll, and Michael~J
  Black.
\newblock Smpl: A skinned multi-person linear model.
\newblock {\em ACM transactions on graphics (TOG)}, 34(6):1--16, 2015.

\bibitem{neuberger2020image}
Assaf Neuberger, Eran Borenstein, Bar Hilleli, Eduard Oks, and Sharon Alpert.
\newblock Image based virtual try-on network from unpaired data.
\newblock In {\em Proceedings of the IEEE/CVF Conference on Computer Vision and
  Pattern Recognition}, pages 5184--5193, 2020.

\bibitem{neverova2018dense}
Natalia Neverova, Riza~Alp Guler, and Iasonas Kokkinos.
\newblock Dense pose transfer.
\newblock In {\em Proceedings of the European conference on computer vision
  (ECCV)}, pages 123--138, 2018.

\bibitem{neverova2020continuous}
Natalia Neverova, David Novotny, Marc Szafraniec, Vasil Khalidov, Patrick
  Labatut, and Andrea Vedaldi.
\newblock Continuous surface embeddings.
\newblock {\em Advances in Neural Information Processing Systems}, 33, 2020.

\bibitem{neverova2019correlated}
Natalia Neverova, David Novotny, and Andrea Vedaldi.
\newblock Correlated uncertainty for learning dense correspondences from noisy
  labels.
\newblock {\em Advances in Neural Information Processing Systems}, 2019.

\bibitem{neverova2019slim}
Natalia Neverova, James Thewlis, Riza~Alp Guler, Iasonas Kokkinos, and Andrea
  Vedaldi.
\newblock Slim densepose: Thrifty learning from sparse annotations and motion
  cues.
\newblock In {\em Proceedings of the IEEE/CVF Conference on Computer Vision and
  Pattern Recognition}, pages 10915--10923, 2019.

\bibitem{nikolenko2019synthetic}
Sergey~I Nikolenko.
\newblock Synthetic data for deep learning.
\newblock {\em arXiv preprint arXiv:1909.11512}, 2019.

\bibitem{pavlakos2019texturepose}
Georgios Pavlakos, Nikos Kolotouros, and Kostas Daniilidis.
\newblock Texturepose: Supervising human mesh estimation with texture
  consistency.
\newblock In {\em Proceedings of the IEEE/CVF International Conference on
  Computer Vision}, pages 803--812, 2019.

\bibitem{rematas2018soccer}
Konstantinos Rematas, Ira Kemelmacher-Shlizerman, Brian Curless, and Steve
  Seitz.
\newblock Soccer on your tabletop.
\newblock In {\em Proceedings of the IEEE Conference on Computer Vision and
  Pattern Recognition}, pages 4738--4747, 2018.

\bibitem{ronneberger2015u}
Olaf Ronneberger, Philipp Fischer, and Thomas Brox.
\newblock U-net: Convolutional networks for biomedical image segmentation.
\newblock In {\em International Conference on Medical image computing and
  computer-assisted intervention}, pages 234--241. Springer, 2015.

\bibitem{sanakoyeu2020transferring}
Artsiom Sanakoyeu, Vasil Khalidov, Maureen~S McCarthy, Andrea Vedaldi, and
  Natalia Neverova.
\newblock Transferring dense pose to proximal animal classes.
\newblock In {\em Proceedings of the IEEE/CVF Conference on Computer Vision and
  Pattern Recognition}, pages 5233--5242, 2020.

\bibitem{sun2019dissecting}
Xiaoxiao Sun and Liang Zheng.
\newblock Dissecting person re-identification from the viewpoint of viewpoint.
\newblock In {\em CVPR}, 2019.

\bibitem{varol17_surreal}
G{\"u}l Varol, Javier Romero, Xavier Martin, Naureen Mahmood, Michael~J. Black,
  Ivan Laptev, and Cordelia Schmid.
\newblock Learning from synthetic humans.
\newblock In {\em CVPR}, 2017.

\bibitem{vaswani2017attention}
Ashish Vaswani, Noam Shazeer, Niki Parmar, Jakob Uszkoreit, Llion Jones,
  Aidan~N Gomez, Lukasz Kaiser, and Illia Polosukhin.
\newblock Attention is all you need.
\newblock {\em arXiv preprint arXiv:1706.03762}, 2017.

\bibitem{weng2019photo}
Chung-Yi Weng, Brian Curless, and Ira Kemelmacher-Shlizerman.
\newblock Photo wake-up: 3d character animation from a single photo.
\newblock In {\em Proceedings of the IEEE/CVF Conference on Computer Vision and
  Pattern Recognition}, pages 5908--5917, 2019.

\bibitem{wu2019m2e}
Zhonghua Wu, Guosheng Lin, Qingyi Tao, and Jianfei Cai.
\newblock M2e-try on net: Fashion from model to everyone.
\newblock In {\em Proceedings of the 27th ACM International Conference on
  Multimedia}, pages 293--301, 2019.

\bibitem{xie2021vton}
Zhenyu Xie, Xujie Zhang, Fuwei Zhao, Haoye Dong, Michael~C Kampffmeyer, Haonan
  Yan, and Xiaodan Liang.
\newblock Was-vton: Warping architecture search for virtual try-on network.
\newblock {\em arXiv preprint arXiv:2108.00386}, 2021.

\bibitem{yang2019parsing}
Lu Yang, Qing Song, Zhihui Wang, and Ming Jiang.
\newblock Parsing r-cnn for instance-level human analysis.
\newblock In {\em Proceedings of the IEEE/CVF Conference on Computer Vision and
  Pattern Recognition}, pages 364--373, 2019.

\bibitem{sfu}
KangKang Yin and Goh~Jing Ying.
\newblock Sfu motion capture database.
\newblock \url{http://mocap.cs.sfu.ca/}.

\end{thebibliography}
}

\end{document}